\documentclass{article}

\PassOptionsToPackage{numbers, compress}{natbib}

\usepackage[square,numbers]{natbib}
\usepackage[final]{neurips_2023}

\usepackage[utf8]{inputenc} 
\usepackage[T1]{fontenc}    
\usepackage{hyperref}       
\usepackage{url}            
\usepackage{booktabs}       
\usepackage{amsfonts}       
\usepackage{nicefrac}       
\usepackage{microtype}      
\usepackage{xcolor}         

\usepackage{amsmath}
\usepackage{amssymb}
\usepackage{mathtools}
\usepackage{amsthm}
\usepackage{comment}
\usepackage{multirow}
\usepackage{multicol}

\usepackage{algorithm}
\usepackage{algpseudocode}

\title{Can Physics Informed Neural Operators Self Improve?}

\author{%
  Ritam Majumdar\\
  TCS Research\\
  \texttt{ritam.majumdar@tcs.com} \\
  \And
  Amey Varhade \\
  IIT Guwahati \\
   \texttt{varhade@alumni.iitg.ac.in} \\
  \AND
  Shirish Karande \\
  TCS Research \\
  \texttt{shirish.karande@tcs.com} \\
   \And
  Lovekesh Vig \\
  TCS Research \\
    \texttt{lovekesh.vig@tcs.com} \\
}

\newcommand{\Method}[1]{\item[\textbf{Method:}] #1}

\begin{document}

\maketitle

\begin{abstract}

Self-training techniques have shown remarkable value across many deep learning models and tasks. However, such techniques remain largely unexplored when considered in the context of learning fast solvers for systems of partial differential equations (Eg: Neural Operators). In this work, we explore the use of self-training for Fourier Neural Operators (FNO). Neural Operators emerged as a data driven technique, however, data from experiments or traditional solvers is not always readily available. Physics Informed Neural Operators (PINO) overcome this constraint by utilizing a physics loss for the training, however the accuracy of PINO trained without data does not match the performance obtained by training with data. In this work we show that self-training can be used to close this gap in performance. We examine canonical examples, namely the 1D-Burgers and 2D-Darcy PDEs, to showcase the efficacy of self-training. Specifically, FNOs, when trained exclusively with physics loss through self-training, approach $1.07\times$ for Burgers and $1.02\times$ for Darcy, compared to FNOs trained with both data and physics loss. Furthermore, we discover that pseudo-labels can be used for self-training without necessarily training to convergence in each iteration. A consequence of this is that we are able to discover self-training schedules  that improve upon the baseline performance of PINO in terms of accuracy as well as time.

\end{abstract}

\section{Introduction}

In recent years, the development of rapid solvers for Partial Differential Equations (PDEs) has garnered substantial interest. Various Neural Operators \cite{li2021fourier,tripura2022wavelet,majumdar2023important,gupta2021multiwavelet} have notably been harnessed to expedite PDE solvers, among which Fourier Neural Operators (FNO) \cite{li2021fourier} have gained significant popularity. These have found applications in a myriad of fields including fluid flow, climate change \cite{pathak2022fourcastnet}, and material modeling \cite{you2022learning}. However, Neural Operators, predominantly trained via data-driven methodologies, often confront challenges related to data availability. Physics-Informed Neural Operators (PINO) \cite{li2021physics}, address this by utilizing an physics-informed loss, based on governing PDEs. Nevertheless, a noticeable disparity in performance is evident when comparing PINOs trained with supervisory loss based on ground truth labels to those trained without.

Self-training, a semi-supervised learning approach, initiates by utilizing a model to generate labels for unlabeled instances. Subsequent to predictions, these instances are amalgamated with the labeled dataset, facilitating further model training. Through repeated iterations, this cycle aims to improve the model's predictive capabilities. Self-training has found utility in varied applications including sentiment analysis \cite{gupta2021unsupervised}, text classification \cite{meng-etal-2020-text}, Natural Language Processing \cite{,acl-2010-acl-2010}, object detection \cite{li2022dtgssod}, medical image classification \cite{https://doi.org/10.1002/mp.16312}, human action recognition \cite{10184205}, facial expression identification \cite{SHABBIR2023104770}, Speech Recognition \cite{baevski2020wav2vec,gheini-etal-2023-joint}, Anomaly Detection \cite{feng2021mist}, as well as Genomics and proteomics \cite{ctx24110331700003408}. Intriguingly, self-training approaches have scarcely been explored to expedite PDE solvers. The closest work which resembles using self-training for PDE solvers is \cite{yan2023stpinn}, wherein the authors use self-training for Physics-informed Neural Networks. More specifically, the authors add the most confident collocation points as pseudo-labels to improve the performance of surrogate PINN for single PDE instance. In this work however, we are performing self-training to generalize over entire function-spaces of PDE tasks rather than a single instance of a PDE, and use PINOs instead of PINNs.   

In this work, we make two important contributions: (1) We show that self-training can be used along with PINO to close the gap in performance between PINO trained with just physics loss v/s the performance obtained by including physics + data loss. We illustrate the efficacy of self-training on the 1D-Burgers and 2D-Darcy PDEs. The performance ratio in terms of the  $L2$ error of PINOs trained in term of just physics loss against PINOs trained with physics + data loss is 39.48  for 1D-Burgers and 1.16 for 2D Darcy. We show that self-training improves this ratio to 1.07 and 1.02. (2) We also explore the utility of self-training and pseudo-labels  in reducing the training time of PINOs. In particular we observed that one can follow an early stopping schedule while training a PINO with physics loss and subsequently with physics + pseudo-labels loss . We observe that some schedules are able to obtain an accuracy greater than PINO while utilizing time lesser than that required to train a PINO to convergence on the basis of physics loss. 

The remainder of the paper is structured as follows: Section 2 outlines the Methodology used for self-training, the results and observations are discussed in Section 3 and we highlight the key conclusions in Section 4.

\section{Methodology}

\begin{algorithm}
\label{self-training}
\caption{Methodology for Self-training of PINO}
\begin{algorithmic}[1]
\Require Parametric PDE $N$, Domain Discretization, Train tasks $T_{\text{train}}$ ($A$)
\Ensure Trained Fourier Neural Operator $G_{\theta_{\text{final}}}$
\Method{}
\State Initialize Fourier Neural Operator $G_{\theta_{\text{1}}}$
\State Train $G_{\theta_{\text{1}}}$ using Physics-loss over $T_{\text{train}}$ until $StoppingCriteria$
\For{i (self-train iterations) = 1 to MAX}
   \State Use $G_{\theta_{\text{i}}}$ to generate data labels for $T_{\text{train}}$ as $D_{\text{labels}}$
   \State Initialize a new Neural Operator $G_{\theta_{\text{i+1}}}$ with $G_{\theta_{\text{i}}}$
   \State Train $G_{\theta_{\text{i+1}}}$ using Physics-loss and Supervised-loss over $D_{\text{labels}}$ until $StoppingCriteria$  
   \If{$|Error(G_{\theta_{\text{i+1}}})-Error(G_{\theta_{\text{i}}})|<\epsilon$}
      \State \textbf{break}
   \EndIf
\EndFor

\State \textbf{return} $G_{\theta_{\text{final}}}$
\end{algorithmic}
\end{algorithm}

Fourier Neural Operators $G_\theta$ aim to learn a mapping between two Banach spaces $G_\theta:A\rightarrow U$, where $A$ typically refers to parameterized initial conditions, boundary conditions, or coefficients of the PDE, while $U$ refers to the final solution of the PDE. Algorithm \ref{self-training} describes the methodology for self-training a physics-informed Neural Operator (PINO). We initialize a Fourier Neural Operator, denoted as $G_{\theta_1}$ and train it on input functions $A$ exclusively using physics-loss. Here, parameterized input functions $A$ refer to our train-tasks $T_{train}$. The physics-loss is defined using the governing equations of the Partial Differential Equation $N$. Leveraging the trained FNO from each prior iteration, we deduce pseudo-labels for our training tasks. These pseudo-labels subsequently function as annotated data points, facilitating supervised fine-tuning in the following iterations. In every iteration, we initialize $G_{\theta_{i+1}}$ using its predecessor $G_{\theta_{i}}$, where $i$ is the current self-train iteration. This FNO is then trained on $T_{\text{train}}$, drawing upon both the physics-loss and a supervised-loss on the aforegenerated pseudo-labels. This iterative loop of pseudo-label inference and operator enhancement is sustained until the absolute difference of two consecutive iterations of FNOs $G_{\theta_{i+1}}$ and $G_{\theta_{i}}$ falls below a threshold $\epsilon$ or maximum number of self-train iterations is completed. Stopping Criteria for training a FNO is met when there is no improvement in the train-loss for 100 consecutive epochs or pre-defined number of epochs are completed. 

\section{Results and Observations}

\subsection{Self-training with PINOs trained to convergence}
\begin{table}
    \centering
    \begin{tabular}{ccccccccccc}
         &&\multicolumn{7}{c}{Self-train Iterations}&\multicolumn{2}{c}{Benchmarks}\\
         && 0 (B0) & 1 & 2 & 3 & 4 & 5 & 6 & B1 & B2\\
        \hline
        \multirow{2}{*}{1D-Burger's}& Data $(1e^{-4})$& $74.22$ & $3.28$ & $2.75$ & $2.19$ & $2.11$ & $2.03$ & ${2.02}$ & ${1.88}$ & $16.26$\\ &Phy. $(1e^{-6})$&8.22&6.96&5.42&4.84&4.00&3.66&{3.83}&${2.65}$&$547$\\
        \hline
        \multirow{2}{*}{2D-Darcy}&Data $(1e^{-2})$& $1.15$ & $1.09$ & $1.07$ & $1.03$ & $1.02$ & $1.01$ & ${1.01}$ & ${0.99}$ & $2.36$\\
    &Phy. $(1e^{-2})$&3.22&1.72&1.45&1.22&1.15&0.98&{0.97}&{0.85}&107\\
        \hline
    \end{tabular}
    \caption{Self-training experiments. Benchmark B1 refers to training PINOs using ground-truth data, while Benchmark B2 refers to training FNOs solely on ground-truth data without physics-loss. Benchmark B0 refers to the FNO trained with just physics-informed loss. Quantities in parenthesis denote the order of the errors. Data refers to test-task L2 errors, while Phy. refers to test-task PDE Mean-squared errors.}
    \label{tab:Data_story}
\end{table}

Let us consider three baselines. B0 (or PINO), FNOs trained with just physics-informed loss, B1, FNOs trained with both physics-informed and actual ground-truth data loss, and B2, FNOs trained solely on ground-truth labels. Table 1 has the performance results for all these baselines and significant performance gap can be observed between training a PINO with ground-truth data and PINO without ground-truth data. For example, in case of 1D Burger's B0 has an L2 error of  $7.42e^{-3}$, while B1 has an error of $1.88e^{-4}$. Infact, the performance of PINO is even worse than the conventional FNO in case of 1D-Burger's i.e. B2, $1.63e^{-3}$. Meanwhile, in case of 2D-Darcy, even though the accuracy of PINO (B0) is greater than that of FNO ($1.15e^{-2}$ v/s $2.36e^{-2}$), it pales significantly in comparison of PINO with data loss ($9.91e^{-3}$).

Table 1 demonstrates the result of self-training a PINO. In this experiment the convergence is not defined in terms of epochs, rather in terms of saturation of reduction in error. We notice that even single iteration of self training can provide significant gains. A self-trained PINO is not only able to easily beat the performance of an FNO but can almost match the performance of PINO trained with data. Our results show that the self-training approach, despite lacking access to true ground-truth, significantly outperforms B2 in both data (Burger’s: 8.61, Darcy: 2.33) and physics-loss (Burger’s: 206, Darcy: 125). Compared to B1, it only exhibits slightly higher errors, specifically [($1.07\times$,$1.02\times$) for data L2 error and ($1.45\times$,$1.15\times$) for  physics-loss MSE for Burger’s and Darcy respectively.

\subsection{Self-training with PINOs not-trained to convergence}
\begin{figure}
    \centering
\includegraphics[width=\textwidth,height=12cm]
{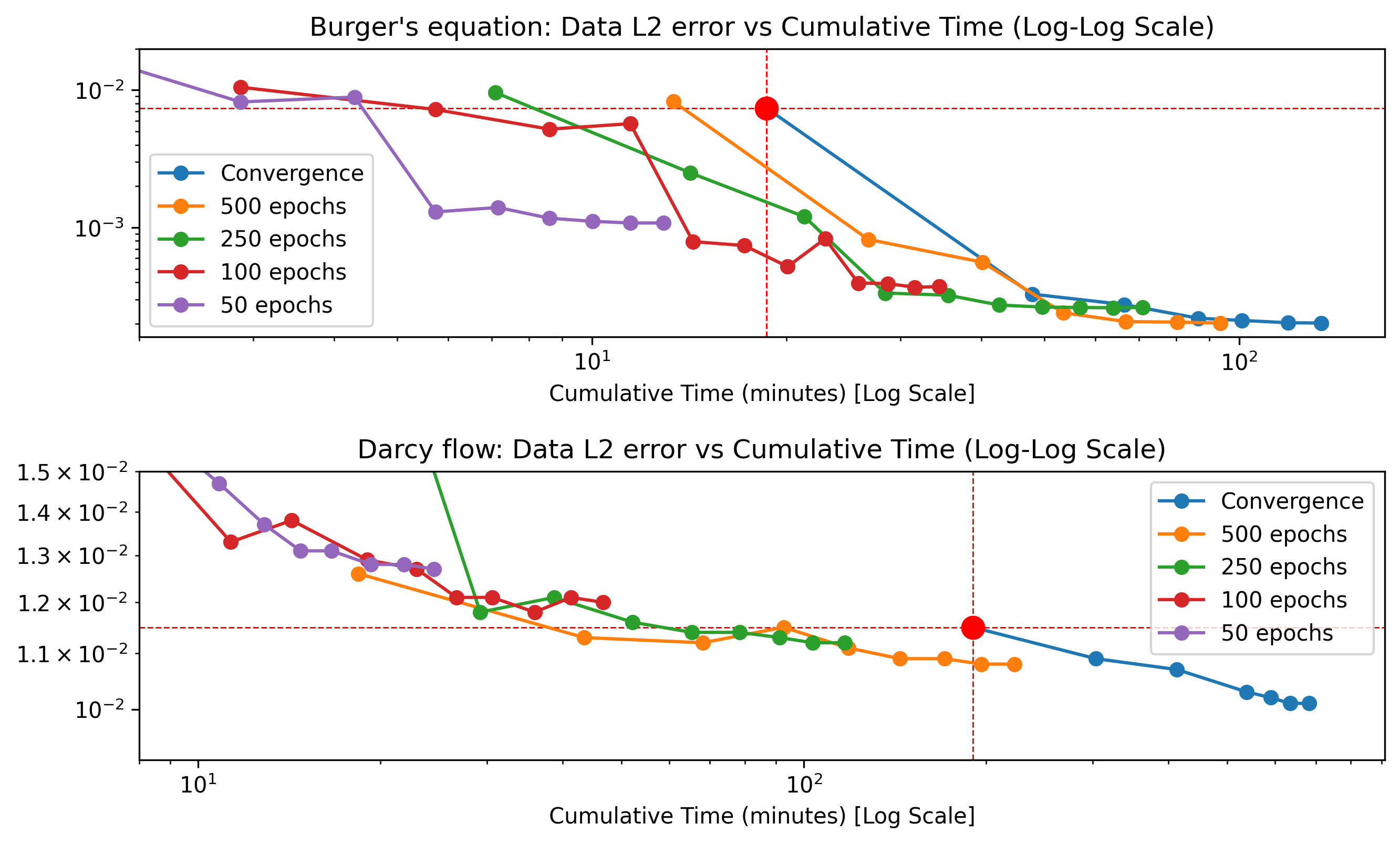}
\caption{Self-training experiments with PINOs trained prematurely. Top: Burger's equation, Bottom: Darcy flow. We study the trend of Data L2-error against cumulative time on a log-log scale. The highlighted point (in red) represents the benchmark B0, i.e. Training a PINO without data-labels.}
    \label{fig:combined}
\end{figure}

Why and when does self-training improve performance? Will models that are not trained to convergence improve more or less with self-training? Poor training will induce noisy pseudo-labels possibly leading to error amplification, however, if the training is just good enough then we might save some epochs of training, without any recognizable loss in inducing pseudo-labels, thus possibly reducing the overall training time. We sought to investigate this possibility. Therefore we trained the models for a fixed number of epochs in each iteration. Indeed a strong self-improvement trend was observed in most cases, until we reach threshold where the models could start diverging. One can observe Figure \ref{fig:combined} closely. In the Burger's equation, PINO without labels has a data L2-error of $7.42e^{-3}$, while taking 18:37 (18 mins and 37 sec) to converge. We observe, self-training for 250 epochs gives us a performance improvement ($2.5e^{-3}$) in the 2nd iteration itself, with a time advantage (14:12), while self-training for 100 epochs on 3rd iteration itself gives us a performance ($5.2e^{-3}$) and time improvement (8:36). Similarly, for Darcy flow, PINO without labels has a data L2-error of $1.15e^{-2}$, while taking (3:10:00) 3 hrs 10 mins to converge. Self-training for 500 epochs for 5 iterations gives us a performance $1.11e^{-2}$ and time improvement (1:58:22), while self-training for 250 epochs at 6th iteration gives us a performance $1.14e^{-2}$ and time improvement (1:05:13). In Figure \ref{fig:combined}, the highlighted point (in red) represents the benchmark B0. All points belonging to the third quadrant with benchmark B0 indicate Pareto-points wherein self-training iterations are both better in performance and time, compared to the benchmark B0. We further provide detailed numerical tabulations of these experiments in the Appendix \ref{tab:Burgers_data_results},\ref{tab:Darcy_data_results}. Thus, we infer that using number of epochs as a hyper-parameter can outperform traditional PINO in terms of error while reducing training time.

\section{Conclusion}
Self-training has been largely unexplored while employing machine learning in numerical methods. In this paper we have successfully shown that self-training with pseudo-label and physics loss can substantially improve the performance of FNOs without the need of any training data. Furthermore we have demonstrated that one can observe the phenomenon of self-improvement through self-training even with noisier FNOs. We have observed that one can stop training early in each iteration of self-training to obtain improvement in accuracy without exceeding the time-budget consumed for training a PINO. We believe that this issue needs further investigation in future.

\section{Broader Impact}

Creation of training data for solvers can consume a lot of compute. We believe this has been one
of the reasons that a lot of data for solved systems does not exist in public domain. This in turn has
meant that the current trend of foundational models which can be used for a broad number of tasks,
has not extended to emergence of instruction tuned foundational models for fast solvers. We believe that if we can train models through
self-improvement, even when they are initially noisy, we can potentially scale the breadth of tasks on
which a single model gets trained. This has the potential to pave a path towards building a foundational
model for a broad set of PDE tasks.

\newpage

\bibliography{neurips_2023.bib}

\begin{thebibliography}{10}

\bibitem{li2021fourier}
Zongyi Li, Nikola Kovachki, Kamyar Azizzadenesheli, Burigede Liu, Kaushik Bhattacharya, Andrew Stuart, and Anima Anandkumar.
\newblock Fourier neural operator for parametric partial differential equations, 2021.

\bibitem{tripura2022wavelet}
Tapas Tripura and Souvik Chakraborty.
\newblock Wavelet neural operator: a neural operator for parametric partial differential equations.
\newblock {\em arXiv preprint arXiv:2205.02191}, 2022.

\bibitem{majumdar2023important}
Ritam Majumdar, Shirish Karande, and Lovekesh Vig.
\newblock How important are specialized transforms in neural operators?
\newblock {\em arXiv preprint arXiv:2308.09293}, 2023.

\bibitem{gupta2021multiwavelet}
Gaurav Gupta, Xiongye Xiao, and Paul Bogdan.
\newblock Multiwavelet-based operator learning for differential equations.
\newblock {\em Advances in neural information processing systems}, 34:24048--24062, 2021.

\bibitem{pathak2022fourcastnet}
Jaideep Pathak, Shashank Subramanian, Peter Harrington, Sanjeev Raja, Ashesh Chattopadhyay, Morteza Mardani, Thorsten Kurth, David Hall, Zongyi Li, Kamyar Azizzadenesheli, et~al.
\newblock Fourcastnet: A global data-driven high-resolution weather model using adaptive fourier neural operators.
\newblock {\em arXiv preprint arXiv:2202.11214}, 2022.

\bibitem{you2022learning}
Huaiqian You, Quinn Zhang, Colton~J Ross, Chung-Hao Lee, and Yue Yu.
\newblock Learning deep implicit fourier neural operators (ifnos) with applications to heterogeneous material modeling.
\newblock {\em Computer Methods in Applied Mechanics and Engineering}, 398:115296, 2022.

\bibitem{li2021physics}
Zongyi Li, Hongkai Zheng, Nikola Kovachki, David Jin, Haoxuan Chen, Burigede Liu, Kamyar Azizzadenesheli, and Anima Anandkumar.
\newblock Physics-informed neural operator for learning partial differential equations.
\newblock {\em arXiv preprint arXiv:2111.03794}, 2021.

\bibitem{gupta2021unsupervised}
Akshat Gupta, Sargam Menghani, Sai~Krishna Rallabandi, and Alan~W Black.
\newblock Unsupervised self-training for sentiment analysis of code-switched data, 2021.

\bibitem{meng-etal-2020-text}
Yu~Meng, Yunyi Zhang, Jiaxin Huang, Chenyan Xiong, Heng Ji, Chao Zhang, and Jiawei Han.
\newblock Text classification using label names only: A language model self-training approach.
\newblock In {\em Proceedings of the 2020 Conference on Empirical Methods in Natural Language Processing (EMNLP)}, pages 9006--9017, Online, November 2020. Association for Computational Linguistics.

\bibitem{acl-2010-acl-2010}
Seniz Demir, Jan Raab, Nils Reiter, Marketa Lopatkova, and Tomek Strzalkowski, editors.
\newblock {\em Proceedings of the {ACL} 2010 Student Research Workshop}, Uppsala, Sweden, July 2010. Association for Computational Linguistics.

\bibitem{li2022dtgssod}
Gang Li, Xiang Li, Yujie Wang, Yichao Wu, Ding Liang, and Shanshan Zhang.
\newblock Dtg-ssod: Dense teacher guidance for semi-supervised object detection, 2022.

\bibitem{https://doi.org/10.1002/mp.16312}
Zhen Peng, Dezhi Zhang, Shengwei Tian, Weidong Wu, Long Yu, Shaofeng Zhou, and Shanhang Huang.
\newblock Faxmatch: Multi-curriculum pseudo-labeling for semi-supervised medical image classification.
\newblock {\em Medical Physics}, 50(5):3210--3222, 2023.

\bibitem{10184205}
Chenxi Wang, Jingzhou Luo, Xing Luo, Haoran Qi, and Zhi Jin.
\newblock V-dixmatch: A semi-supervised learning method for human action recognition in night video sensing.
\newblock {\em IEEE Sensors Journal}, pages 1--1, 2023.

\bibitem{SHABBIR2023104770}
Nazir Shabbir and Ranjeet~Kumar Rout.
\newblock Fgbcnn: A unified bilinear architecture for learning a fine-grained feature representation in facial expression recognition.
\newblock {\em Image and Vision Computing}, 137:104770, 2023.

\bibitem{baevski2020wav2vec}
Alexei Baevski, Henry Zhou, Abdelrahman Mohamed, and Michael Auli.
\newblock wav2vec 2.0: A framework for self-supervised learning of speech representations, 2020.

\bibitem{gheini-etal-2023-joint}
Mozhdeh Gheini, Tatiana Likhomanenko, Matthias Sperber, and Hendra Setiawan.
\newblock Joint speech transcription and translation: Pseudo-labeling with out-of-distribution data.
\newblock In {\em Findings of the Association for Computational Linguistics: ACL 2023}, pages 7637--7650, Toronto, Canada, July 2023. Association for Computational Linguistics.

\bibitem{feng2021mist}
Jia-Chang Feng, Fa-Ting Hong, and Wei-Shi Zheng.
\newblock Mist: Multiple instance self-training framework for video anomaly detection, 2021.

\bibitem{ctx24110331700003408}
Zhuohan Yu, Yanchi Su, Yifu Lu, Yuning Yang, Fuzhou Wang, Shixiong Zhang, Yi~Chang, Ka-Chun Wong, and Xiangtao Li.
\newblock Topological identification and interpretation for single-cell gene regulation elucidation across multiple platforms using scmgca.
\newblock {\em Nature communications.}, 14(1), 2023-12-25.

\bibitem{yan2023stpinn}
Junjun Yan, Xinhai Chen, Zhichao Wang, Enqiang Zhoui, and Jie Liu.
\newblock St-pinn: A self-training physics-informed neural network for partial differential equations, 2023.

\end{thebibliography}

\appendix
\section{Appendix}
\subsection{PDE information}

\subsubsection{Burgers' Equation} 

We consider the 1D Burgers' equation, a non-linear PDE with periodic boundary conditions. The governing equations are defined as follows:
$$
\begin{aligned}
\partial_t u(x, t)+\partial_x\left(u^2(x, t) / 2\right) & =\nu \partial_{x x} u(x, t), & & x\in(0,1), t\in(0,1] \\
u(x, 0) & =u_0(x), & & x\in(0,1)
\end{aligned}
$$
Here $u_0$ is the initial condition and $\nu = 1e^{-2}$ is the viscosity coefficient. We aim to learn the operator mapping the initial condition $u_0 $ to the entire solution at $u|_{[0,1]}$. 

\subsubsection{2D Darcy Flow} 2D Darcy Flow is a linear steady-state second-order elliptic PDE. We consider a flow in a unit box, whose governing equations are given by:
$$
\begin{aligned}
-\nabla \cdot(a(x) \nabla u(x)) & =f(x), & & x \in(0,1)^2 \\
u(x) & =0, & & x \in \partial\:(0,1)^2
\end{aligned}
$$
Here $a$ is the piecewise constant diffusion coefficient and $f=1$ is the forcing function. The objective is to learn the operator mapping the diffusion coefficient $a(x)$ to the solution $u(x)$. 

\subsection{Training and Hyperparameter Details}
\label{training_and_hyperparameter_details}

As our base architecture, we consider a FNO stacked with 4 layers with 15,12,9,9 fourier modes in every layer, using GeLU as our activation function to model non-linearities. We consider 800 train and 200 test examples in Burger's equation, while we consider 1000 train and 500 test examples in Darcy equation respectively. The prediction gridsize of $[x,t]\rightarrow[128,101]$ for Burger's equation, while for Darcy equation it's $[x,y]\rightarrow[61,61]$. All experiments were conducted on Nvidia P100 GPU with 16 GB GPU Memory and 1.32 GHz GPU Memory clock using Pytorch framework. In our benchmark experiments B1, B2 and for self-training experiments trained till convergence, we start with a learning rate of $1e^{-3}$, with a patience of 100 and tolerance of $1e^{-5}$ for the total train error, until the minimum learning rate of $1e^{-6}$ is obtained. In the self-training experiments wherein the Neural Operators are trained for 250, 100 and 50 epochs (not till convergence): We start with a base learning rate of $1e^{-3}$, and reduce the learning rate by a factor of 0.1 when there is no-improvement in the test error at start of every iteration. The physics-informed loss is evaluated using Finite Difference Method in the spectral domain. In benchmark experiments B0 and B1, the dataset is generated using Finite-Difference Method in the spectral domain with input conditions generated using Gaussian Random Fields.

\subsection{Tabulated Results of Self-training experiments}

\begin{table}[h]
    \caption{1D Burger's equation results}
    \centering
    \begin{tabular}{c|c|c|c|c|c|c}
\hline
Iteration&&Conv.&500 epochs&250 epochs&100 epochs&50 epochs\\
\hline
\vspace{1pt}
&Data L2-error&7.42$e^{-3}$&8.26$e^{-3}$&9.65$e^{-3}$&1.05$e^{-2}$&2.23$e^{-2}$\\
0&Physics MSE&8.22$e^{-6}$&9.81$e^{-6}$&5.33$e^{-5}$&7.48$e^{-5}$&6.58$e^{-4}$\\
&Cum. time&18:37&13:22&7:06&2:52&1:26\\
\hline
&Data L2-error&3.28$e^{-4}$&8.25$e^{-4}$&2.56$e^{-3}$&7.24$e^{-3}$&8.26$e^{-3}$\\
1&Physics MSE&6.96$e^{-6}$&1.17$e^{-5}$&1.06$e^{-5}$&3.06$e^{-5}$&8.01$e^{-4}$\\
&Cum. time&47:53&26:44&14:12&5:44&2:52\\
\hline
&Data L2-error&2.75$e^{-4}$&5.69$e^{-4}$&1.28$e^{-3}$&5.21$e^{-3}$&8.91$e^{-3}$\\
2&Physics MSE&5.42$e^{-6}$&7.62$e^{-6}$&6.97$e^{-6}$&3.57$e^{-5}$&7.29$e^{-4}$\\
&Cum. time&1:06:30&40:06&21:18&8:36&4:18\\
\hline
&Data L2-error&2.19$e^{-4}$&2.45$e^{-4}$&3.34$e^{-4}$&5.73$e^{-4}$&1.32$e^{-3}$\\
3&Physics MSE&4.84$e^{-6}$&4.66$e^{-6}$&5.01$e^{-6}$&3.85$e^{-5}$&4.10$e^{-4}$\\
&Cum. time&1:26:24&53:28&28:24&11:28&5:44\\
\hline
&Data L2-error&2.11$e^{-4}$&2.07$e^{-4}$&3.22$e^{-4}$&7.92$e^{-4}$&1.42$e^{-3}$\\
4&Physics MSE&4.00$e^{-6}$&3.90$e^{-6}$&4.43$e^{-6}$&1.05$e^{-5}$&4.10$e^{-4}$\\
&Cum. time&1:41:06&1:06:50&35:30&14:20&7:10\\
\hline
&Data L2-error&2.03$e^{-4}$&2.05$e^{-4}$&2.73$e^{-4}$&7.42$e^{-4}$&1.17$e^{-3}$\\
5&Physics MSE&3.66$e^{-6}$&3.85$e^{-6}$&4.16$e^{-6}$&8.95$e^{-6}$&4.05$e^{-4}$\\
&Cum. time&1:59:00&1:20:12&42:35&17:12&8:36\\
\hline
&Data L2-error&2.02$e^{-4}$&2.02$e^{-4}$&2.64$e^{-4}$&5.25$e^{-4}$&1.15$e^{-3}$\\
6&Physics MSE&3.83$e^{-6}$&3.83$e^{-6}$&3.91$e^{-6}$&8.15$e^{-6}$&3.97$e^{-4}$\\
&Cum. time&2:13:49&1:33:34&49:40&20:04&10:02\\
\hline
&Data L2-error&&&2.62$e^{-4}$&8.36$e^{-4}$&1.08$e^{-3}$\\
7&Physics MSE&&&3.84$e^{-6}$&7.55$e^{-6}$&3.97$e^{-4}$\\
&Cum. time&&&56:45&22:56&11:28\\
\hline
&Data L2-error&&&2.61$e^{-4}$&3.95$e^{-4}$&1.08$e^{-3}$\\
8&Physics MSE&&&3.82$e^{-6}$&7.39$e^{-6}$&3.96$e^{-4}$\\
&Cum. time&&&1:03:50&25:48&12:54\\
\hline
&Data L2-error&&&2.62$e^{-4}$&3.91$e^{-4}$&\\
9&Physics MSE&&&3.81$e^{-6}$&7.39$e^{-6}$&\\
&Cum. time&&&1:10:55&28:40&\\
\hline
&Data L2-error&&&&3.68$e^{-4}$&\\
10&Physics MSE&&&&7.39$e^{-6}$&\\
&Cum. time&&&&31:32&\\
\hline
&Data L2-error&&&&3.72$e^{-4}$&\\  
11&Physics MSE&&&&7.38$e^{-6}$&\\
&Cum. time&&&&34:24&\\
\hline
\end{tabular}
    \label{tab:Burgers_data_results}
\end{table}

\begin{table}[h]
    \centering
    \caption{2D Darcy flow results}
    \begin{tabular}{c|c|c|c|c|c|c}
\hline
Iteration&&Conv.&500 epochs&250 epochs&100 epochs&50 epochs\\
\hline

\vspace{1pt}
&Data L2-error&1.15$e^{-2}$&1.26$e^{-2}$&1.63$e^{-2}$&3.22$e^{-2}$&6.96$e^{-2}$\\
0&Physics MSE&3.22$e^{-2}$&5.46$e^{-2}$&8.71$e^{-2}$&1.21$e^{-1}$&1.72$e^{-1}$\\
&Cum. time&3:10:00&18:24&9:41&4:02&2:13\\
\hline

&Data L2-error&1.09$e^{-2}$&1.13$e^{-2}$&2.19$e^{-2}$&1.61$e^{-2}$&2.92$e^{-2}$\\
1&Physics MSE&1.72$e^{-2}$&2.20$e^{-2}$&7.78$e^{-2}$&1.02$e^{-1}$&1.35$e^{-1}$\\
&Cum. time&5:03:37&45:22&18:38&7:46&4:08\\
\hline

&Data L2-error&1.07$e^{-2}$&1.12$e^{-2}$&1.18$e^{-2}$&1.33$e^{-2}$&3.18$e^{-2}$\\
2&Physics MSE&1.45$e^{-2}$&1.84$e^{-2}$&3.41$e^{-2}$&7.98$e^{-2}$&1.21$e^{-1}$\\
&Cum. time&6:52:59&1:10:06&29:18&11:21&6:01\\
\hline

&Data L2-error&1.03$e^{-2}$&1.15$e^{-2}$&1.21$e^{-2}$&1.38$e^{-2}$&1.64$e^{-2}$\\
3&Physics MSE&1.22$e^{-2}$&1.60$e^{-2}$&2.91$e^{-2}$&7.76$e^{-2}$&5.51$e^{-2}$\\
&Cum. time&8:57:20&1:34:37&38:46&14:18&8:39\\
\hline

&Data L2-error&1.02$e^{-2}$&1.11$e^{-2}$&1.16$e^{-2}$&1.29$e^{-2}$&1.47$e^{-2}$\\
4&Physics MSE&1.15$e^{-2}$&1.44$e^{-2}$&1.88$e^{-2}$&3.36$e^{-2}$&4.72$e^{-2}$\\
&Cum. time&9:48:42&2:00:22&52:13&18:42&10:49\\
\hline

&Data L2-error&1.01$e^{-2}$&1.09$e^{-2}$&1.14$e^{-2}$&1.27$e^{-2}$&1.37$e^{-2}$\\
5&Physics MSE&9.81$e^{-3}$&1.32$e^{-2}$&1.72$e^{-2}$&3.26$e^{-2}$&4.63$e^{-2}$\\
&Cum. time&10:33:56&2:25:56&1:05:23&22:38&12:51\\
\hline

&Data L2-error&1.01$e^{-2}$&1.09$e^{-2}$&1.14$e^{-2}$&1.21$e^{-2}$&1.31$e^{-2}$\\
6&Physics MSE&9.79$e^{-3}$&1.30$e^{-2}$&1.63$e^{-2}$&3.15$e^{-2}$&4.85$e^{-2}$\\
&Cum. time&11:22:17&2:52:25&1:18:21&26:23&14:45\\
\hline

&Data L2-error&&1.08$e^{-2}$&1.13$e^{-2}$&1.21$e^{-2}$&1.31$e^{-2}$\\
7&Physics MSE&&1.28$e^{-2}$&1.56$e^{-2}$&2.93$e^{-2}$&4.46$e^{-2}$\\
&Cum. time&&3:18:54&1:31:18&30:28&16:35\\
\hline

&Data L2-error&&1.08$e^{-2}$&1.12$e^{-2}$&1.18$e^{-2}$&1.28$e^{-2}$\\
8&Physics MSE&&1.27$e^{-2}$&1.51$e^{-2}$&2.14$e^{-2}$&3.27$e^{-2}$\\
&Cum. time&&3:45:16&1:43:31&35:47&19:17\\
\hline

&Data L2-error&&&1.12$e^{-2}$&1.21$e^{-2}$&1.28$e^{-2}$\\
9&Physics MSE&&&1.36$e^{-2}$&1.99$e^{-2}$&3.12$e^{-2}$\\
&Cum. time&&&1:56:34&41:06&21:51\\
\hline

&Data L2-error&&&&1.23$e^{-2}$&1.27$e^{-2}$\\
10&Physics MSE&&&&1.89$e^{-2}$&3.09$e^{-2}$\\
&Cum. time&&&&46:25&24:30\\

\hline
\end{tabular}
    \label{tab:Darcy_data_results}
\end{table}

\end{document}